\documentclass[10pt,twocolumn,letterpaper]{article}

\usepackage{iccv}              

\usepackage{tabularx}    
\usepackage{multirow}    
\usepackage{booktabs}    
\usepackage{float}
\usepackage{array}       
\usepackage{bm}
\usepackage{overpic}
\newcommand{\ourMethod}[0]{\textsc{PRISM}}

\definecolor{iccvblue}{rgb}{0.21,0.49,0.74}
\usepackage[pagebackref,breaklinks,colorlinks,allcolors=iccvblue]{hyperref}

\title{PRISM: Probabilistic Representation for Integrated Shape\\ Modeling and Generation}

\author{Lei Cheng$^{1}$ \quad 
Mahdi Saleh$^{1}$ \quad
Qing Cheng$^{1}$ \quad \\
Lu Sang$^{1}$ \quad 
Hongli Xu$^{1}$ \quad 
Daniel Cremers$^{1}$ \quad 
Federico Tombari$^{1,2}$\\
$^1$ Technical University of Munich \quad $^2$ Google}

\begin{document}

\twocolumn[{%
\renewcommand\twocolumn[1][]{#1}%
\maketitle
\vspace{-35pt}
\begin{center}
\centering
\captionsetup{type=figure}
\includegraphics[width=\linewidth]{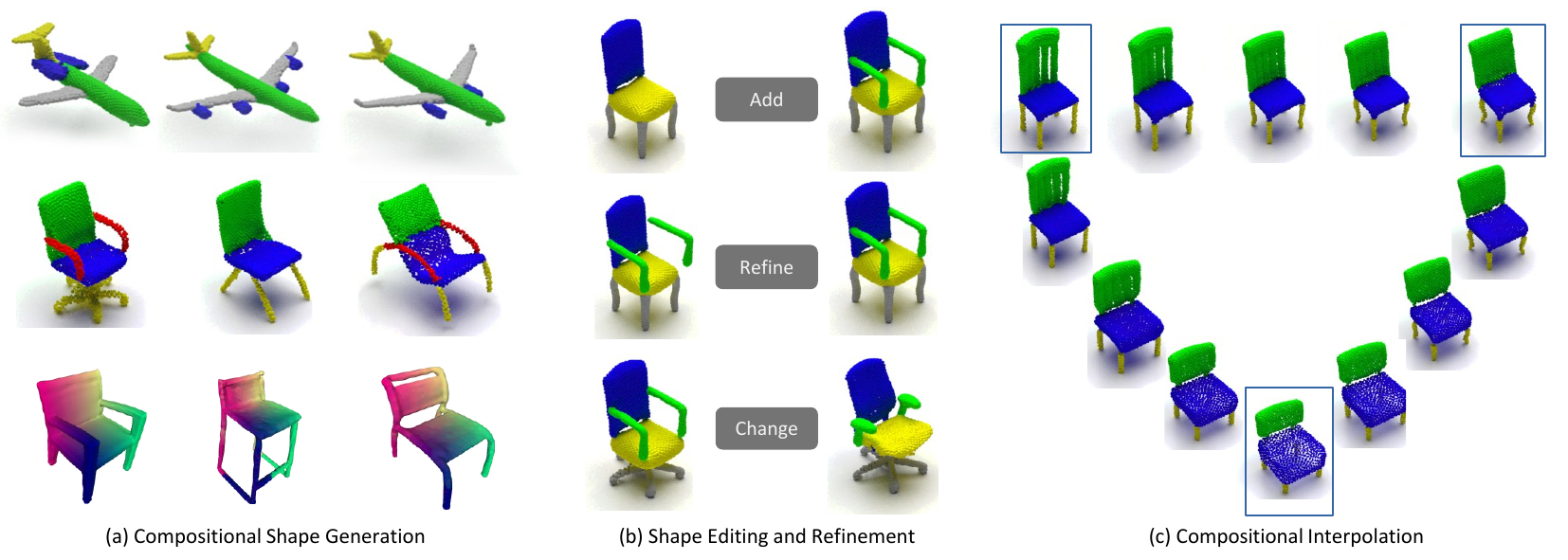}
\vspace{-10pt}
\caption{PRISM: A novel framework for 3D shape modeling that enables (a) diverse compositional shape generation and accurate texture mapping, (b) intuitive shape editing and refinement operations including adding, refining, and changing parts, and (c) smooth compositional interpolation between different shape variations while preserving structural coherence.}
\label{fig:teaser}
\end{center}
}]

\begin{abstract}
Despite the advancements in 3D full-shape generation, accurately modeling complex geometries and semantics of shape parts remains a significant challenge, particularly for shapes with varying numbers of parts. Current methods struggle to effectively integrate the contextual and structural information of 3D shapes into their generative processes. We address these limitations with PRISM, a novel compositional approach for 3D shape generation that integrates categorical diffusion models with Statistical Shape Models (SSM) and Gaussian Mixture Models (GMM). Our method employs compositional SSMs to capture part-level geometric variations and uses GMM to represent part semantics in a continuous space. This integration enables both high fidelity and diversity in generated shapes while preserving structural coherence. Through extensive experiments on shape generation and manipulation tasks, we demonstrate that our approach significantly outperforms previous methods in both quality and controllability of part-level operations. Our code will be made publicly available.\footnote{Repository link.}
\end{abstract}    
\section{Introduction}
\label{sec:intro}

Understanding and representing three-dimensional objects in their compositional nature remains a fundamental challenge in computer vision and graphics. While recent advances in deep learning have revolutionized geometric modeling, real-world objects are naturally organized into meaningful components with specific relationships and variations beyond a single unique shape. This compositional understanding underpins crucial applications across domains: from computer-aided design, where objects are constructed from functional components, to robotic manipulation, where understanding part relationships enables physical interaction, to medical imaging, where anatomical structures follow specific compositional patterns. Yet current approaches often treat objects as purely geometric entities, missing the rich structural patterns that govern how real-world objects are composed and function.

A key challenge in modeling real-world objects is their structural variability and complexity. Unlike simplified 3D models, everyday objects exhibit varying part configurations where components may be present or absent \cite{nakayama2023difffacto, koo2023salad} and must follow valid compositional rules. While recent works have made progress in geometric accuracy \cite{gao2022get3d} or part-level modeling \cite{mo2019structurenet}, effectively representing both shape variations and valid compositional patterns remains difficult. Recent full-shape generation methods \cite{gao2022get3d, cai2020learning} lack explicit structural control, while part-based approaches \cite{nakayama2023difffacto, koo2023salad} lack statistical understanding of shape variations. This gap necessitates capturing the geometric details of individual parts, understanding their semantic relationships, and ensuring structural validity - a challenge that current approaches have yet to address in a unified framework.

Statistical Shape Models (SSMs) have proven highly successful in domains with consistent topology, providing a principled approach to modeling shape variations. In medical imaging, SSMs effectively capture anatomical variations \cite{heimann2009statistical}. At the same time, in human modeling, methods like SMPL \cite{loper2023smpl} for bodies, MANO \cite{romero2022embodied} for hands, and FLAME \cite{li2017learning} for faces have become standard tools due to their compact representation and statistical guarantees. These models excel when point correspondences are well-defined, and topology remains fixed. However, applying similar statistical approaches to everyday objects poses significant challenges due to their varying topology and lack of point-wise correspondence. While recent works have attempted to extend SSMs to more general shapes \cite{adams2023point2ssm, bastian2023s3m}, they still struggle with objects that exhibit structural variations.

To address these challenges, we propose a novel integration of SSMs with categorical diffusion for compositional shape generation. Our key insight is to combine part-level statistical modeling through SSMs with categorical diffusion to handle varying part configurations while learning correspondences in an unsupervised manner. This integration allows us to capture both geometric variations and valid compositional patterns without manual annotations, providing better morphological control through an interpretable, statistically grounded framework.

Our key technical innovations lie in two aspects: a compact part-wise SSM representation and a tailored categorical diffusion process. This combination enables our model to learn and generate shapes with varying topology, effectively capturing the statistical patterns of how parts appear, vary, and combine in real-world objects, directly addressing previous approaches' limitations.
In summary, our main contributions are:
\begin{itemize}
\item A novel compostional object representation combining part-wise semantics and SSMs that captures local geometric and topological variations.
\item An unsupervised approach to learning part-level statistical shape variations without manual annotations, making our method scalable to real-world object categories.
\item A novel structured generative process with categorical diffusion that effectively hands the combinatorial nature of real-world object compositions.
\item Comprehensive evaluation on real-world datasets shows superior generation quality and manipulation control performance, particularly for objects with varying structures.
\end{itemize}
\section{Related Work}

\subsection{3D Shape Generation Methods}
Recent advancements in 3D shape generation have revolutionized how we create and manipulate three-dimensional objects. Methods such as GET3D \cite{gao2022get3d} and ShapeGF \cite{cai2020learning} have impressive capabilities in generating high-quality 3D shapes with realistic textures. However, these full-shape generation approaches treat objects as monolithic entities, lacking explicit control over their structural composition. PointFlow \cite{yang2019pointflow} pioneered the use of normalizing flows for point cloud generation, while StructureNet \cite{mo2019structurenet} introduced hierarchical graph networks to capture structural relationships. Recent approaches like MeshGPT \cite{siddiqui2024meshgpt}, PolyGen \cite{nash2020polygen} and Polydiff \cite{alliegro2023polydiff} have focused on directly generating triangle meshes using transformer architectures or diffusion models. Despite these advances, there remains to be a challenge to effectively represent shapes and valid compositional patterns that characterize real-world objects.

\subsection{Part-based and Compositional Modeling}
Part-based approaches offer more fine-grained control over shape generation and manipulation. DiffFacto \cite{nakayama2023difffacto} introduced a controllable part-based generation framework using cross-diffusion, while SALAD \cite{koo2023salad} proposed a part-level latent diffusion model for shape manipulation. CompoNet \cite{schor2019componet} and PQ-NET \cite{wu2020pq} explored part synthesis and composition for generating novel shapes. AutoSDF \cite{mittal2022autosdf} leveraged shape priors for completion and generation tasks. SAGNet \cite{wu2019sagnet} uses a VAE to encode part geometry and pairwise relations, while GRASS \cite{li2017grass} models part structure hierarchies as binary trees. SDM-NET \cite{gao2019sdm} combines deformable mesh boxes to represent part geometry and a VAE to encode global structure, and DSG-Net \cite{yang2020dsm} disentangles part geometry from structure by learning separate latent spaces. However, these approaches often lack a statistical understanding of shape variations, limiting their ability to capture valid compositional patterns and morphological control. Our work addresses this limitation by integrating part-level statistical modeling with categorical diffusion to handle varying part configurations while maintaining statistical guarantees.

\subsection{SSMs and Learning Correspondences}
Statistical modeling have long been widely used for representing shape variations in domains with consistent topology. In medical imaging, SSMs have been extensively used to model anatomical variations \cite{heimann2009statistical}, while in human modeling, methods like SMPL \cite{loper2023smpl}, MANO \cite{romero2022embodied}, and FLAME \cite{li2017learning} have become standard tools for bodies, hands, and faces, respectively. Recent works such as Point2SSM \cite{adams2023point2ssm} and S3M \cite{bastian2023s3m} have attempted to extend SSMs to more general shapes, but they still struggle with objects that exhibit structural variations. DeepSDF \cite{park2019deepsdf} and structured implicit functions \cite{genova2019learning} have explored learning-based approaches to shape representation without explicit part-level statistical modeling.

\subsection{Articulated Object Modeling}
Recent advancements in articulated object modeling have focused on reconstructing or generating objects with geometric and motion properties. Ditto \cite{jiang2022ditto} reconstructs articulated objects from point clouds as implicit representations, capturing geometric occupancy, part types, and joint parameters. PARIS \cite{liu2023paris} disentangles static and movable parts from observations at different articulation states. NAP \cite{lei2023nap} introduces the unconditional generation of articulated objects by modeling parameters with geometry using an articulation tree. CAGE \cite{liu2024cage} employs diffusion to generate articulation abstractions, relying on part retrieval for final object geometry. However, these methods focus on reconstruction rather than generation or rely on part retrieval, which can introduce inconsistencies in part geometry and overall shape structure. Our approach uniquely combines the strengths of SSMs with categorical diffusion to handle the compositional nature of objects with varying topology.
\section{PRISM }
\label{sec:method}

\subsection{Background on Diffusion Model}
\label{sec:background}
Diffusion models have emerged as powerful generative frameworks capable of producing high-quality samples across various domains. In this section, we review key concepts of diffusion models underpinning our approach, including continuous and discrete formulations.


\noindent \textbf{Continuous diffusion models.} As a continuous latent variable model~\cite{ho2020denoising}, a diffusion (i.e. forward) process is modeled as a Markov noising process with $T$ steps $\left\{\mathbf{x}_t\right\}_{t=0}^T$, where $\mathbf{x}_0 \sim q\left(\mathbf{x}_0\right)$ is sampled from the target data distribution and $\mathbf{x}^{(T)} \sim  \mathcal{N} (\mathbf{0}, \mathbf{I}) $ is the standard normal prior. The data $\mathbf{x}_0$ is progressively noised through a diffusion process, transforming it into latent variables within the same sample space, $q\left(\mathbf{x}^{(1: t)} \mid \mathbf{x}^{(0)}\right):=\Pi_{s=1}^t q\left(\mathbf{x}^{(s)} \mid \mathbf{x}^{(s-1)}\right)$
, where $q\left(\mathbf{x}^{(s)} \mid \mathbf{x}^{(s-1)}\right):=\mathcal{N}\left(\mathbf{x}^{(s)} ; \sqrt{1-\beta^{(s)}} \mathbf{x}^{(s-1)}, \beta^{(s)} \mathbf{I}\right)$, and ${\beta^{(s)}}$ is a predefined Gaussian noise variance schedule. Therefore, a noised data in forward process, $\mathbf{x}^{(t)} = \sqrt{\bar{\alpha}^{(t)}} \mathbf{x}^{(0)}+\sqrt{1-\bar{\alpha}^{(t)}} \epsilon $ , $\alpha^{(t)}:=1-\beta^{(t)}$, $\bar{\alpha}^{(t)}:=\Pi_{s=1}^t \alpha^{(s)}$, and $\epsilon $ is sampled from $\mathcal{N} (\mathbf{0}, \mathbf{I})$ .

The corresponding reverse denoising process from $\mathbf{x}_T$ to  $\mathbf{x}_0$  is defined as a parameterized Markov chain: 
\begin{equation}
\begin{aligned}
p_\theta\left(\mathbf{x}^{(0)}\right):=\int p_\theta\left(\mathbf{x}^{(0: T)}\right) d \mathbf{x}^{(1: T)} , \\ 
p_{\theta} (\mathbf{x}^{(0:T)}) = p (\mathbf{x}^{(T)}) \Pi_{t=1}^{T} p_{\theta} (\mathbf{x}^{(t-1)} \vert \mathbf{x}^{(t)}).
\end{aligned}
\end{equation}
In the variational inference framework~\cite{blei2017variational}, we optimize network weights $\theta$  by optimizing the following variational bound on the negative log-likelihood:
\begin{equation}
\begin{aligned}
    \vspace{-5pt}
    \mathbb{E}_{q (\mathbf{x}^{(0)})} & [-\log p_{\theta} (\mathbf{x}^{(0)})] \leq \\
    &\mathbb{E}_{q (\mathbf{x}^{(0)}, \dots, \mathbf{x}^{(T)})} \left[-\log \frac{p_{\theta} (\mathbf{x}^{(0:T)})}{q (\mathbf{x}^{(1:T)} \vert \mathbf{x}^{(0)})}\right].
\end{aligned}
\end{equation}

Following Ho et al.\cite{ho2020denoising}, the conditional probabilities,
$p_\theta\left(\mathbf{x}^{(t-1)} \mid \mathbf{x}^{(t)}\right):=\mathcal{N}\left(\mathbf{x}^{(t-1)} ; \boldsymbol{\mu}_\theta\left(\mathbf{x}^{(t)}, t\right), \beta^{(t)} \mathbf{I}\right) $.

We predict $\mathbf{x}^{(0)}$ directly for the target latent with a training objective:
\begin{equation}
\mathcal{L}(\theta):=\mathbb{E}_{t, \mathbf{x}^{(0)}, \boldsymbol{\epsilon}}\left[\left\|\mathbf{x}^{(0)}-\boldsymbol{\epsilon}_\theta\left(\mathbf{x}^{(t)}, t\right)\right\|^2\right]
\end{equation}



\noindent \textbf{Discrete diffusion models.} To generalize the diffusion process to discrete cases, Austin et al.~\cite{austin2021structured} propose applying a Markov noising process directly to the probability vector representing categorical distributions. Specifically, considering discrete random variables with $K$ distinct categories, denoted as $y_0, \dots, y_T \in {1, \dots, K}$, the initial state $y_0$ is sampled from the underlying data distribution. The forward diffusion (noising) process is defined through a Markov transition matrix:
$[Q_t]{i,j} = q(y_t = i \mid y_{t-1} = j).$
When diffusing a categorical dataset with components $\mathbf{y}_0 \in {0, \dots, K}^N$, the state transition matrix $Q_t$ is applied to each state vector across categorical elements, defining $q(\mathbf{x}_t \mid \mathbf{x}_{t-1}) = \operatorname{Cat}(\mathbf{x}_t; \mathbf{p} = \mathbf{x}_{t-1} Q_t)$
where \quad $\bar{Q}_t = Q_1 Q_2 \dots Q_t$.
This process effectively governs transitions between categorical states or their absorption into noise states. To reverse this diffusion process, a denoising network is trained to predict class probabilities directly by minimizing a cross-entropy loss.

\begin{figure*}[t]
\centering
\includegraphics[width=0.93\linewidth]{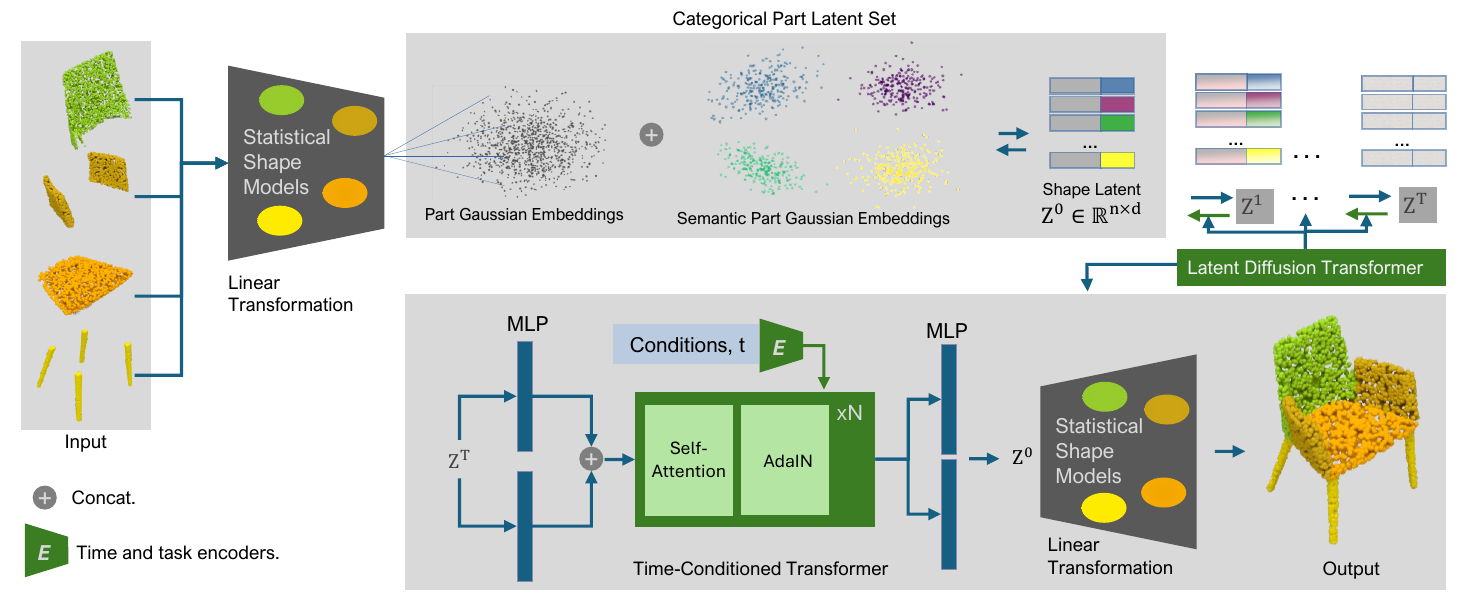}
\caption{\textbf{Method Overview.} 
Given a segmented 3D shape $S$ of $m$ parts, we encode the part point cloud with Statistical shape models and represent it as an unordered set. Each part set has its own SSM, the encoder and decoder are a group of the same categorical SSMs. We represent each categorical part with Gaussian embedding and Gaussian semantics embedding from the Gaussian Mixture Model. Then we train our part-level latent diffusion model on the part sets for conditional/unconditional 3D shape generation. The generated latent part set can be decoded and denoised into a clean 3D shape.}
\label{fig:overview}
\end{figure*}

\subsection{Problem Overview} Our method aims to learn a part-level generative model for 3D shape generation compositionally.
Given a set of segmented 3D shapes $\mathbf{S}$ consisting of part sets $ \mathbf{V}$ and their semantic categories $\mathbf{C}$, we want to decompose the target of learning distribution $P(\mathbf{S})$ to learning the conditional distribution $P(\mathbf{S} \mid \mathbf{V}, \mathbf{C})$ and the joint distribution $P(\mathbf{V}, \mathbf{C})$ according to 
$P(\mathbf{S}) = P(\mathbf{S} \mid \mathbf{V}, \mathbf{C}) P(\mathbf{V}, \mathbf{C})
$. 

Unlike previous approaches that parameterize each part instance and learn separate distributions~\cite{nakayama2023difffacto}, we first model the part to full shape mapping, $P(\mathbf{S} \mid \mathbf{V}, \mathbf{C})$, with our proposed data representation in section~\ref{sec:part_level_rep} and then learn the distribution of $P(\mathbf{V}, \mathbf{C})$ with a categorical diffusion model in section~\ref{sec:cate_part_set_diffusion}. The overview is illustrated in Fig.~\ref{fig:overview}.

In this way, we can capture the statistical distribution of part-level shape variations within each semantic category and also the global shape distribution, including the geometry style and part topology. 


\subsection{Part-Level Shape Representation}
\label{sec:part_level_rep}


\noindent \textbf{Notation.} Given a set of segmented 3D shapes $\mathbf{S}=\{S^{(i)}\}_{i \in \{1, \ldots, n\}}$, a shape $S^{(i)}$ can be decomposed into semantic parts $\{S_j^{(i)}\}_{j \in \{1, \ldots, m\}}$, where $m$ denotes the maximal number of semantic parts of the whole set $\mathbf{S}$. When grouping parts by semantic category, we form a part set $ \mathbf{V}_j=\{S_j^{(i)}\} $ for each semantic part, e.g. chair legs. Thus, the shape dataset can be represented as $\mathbf{S}=\cup_{j=1}^m \mathbf{V}_j$. We use $\mathbf{C}= \{c_1, c_2, ...c_m \}$ to denote the part semantic labels.

\noindent \textbf{Part-Level SSM}. To parameterize the part-level shape variations, we pre-compute linear statistical shape models (SSM) based on the principle component analysis (PCA) for each shape part set. 
Inspired by point2SSM~\cite{adams2023point2ssm}, we first sample a fixed number of points from the part meshes and then establish one-to-one point correspondence among all part points. This makes it inherently easier to capture the part-level similarities than the entire shape structure. 


We use the linear SSM to encode the shape part point clouds into latent samples following multi-variant Gaussian distribution. Given correspondence points $ \mathbf{X} = \left\{x_i\right\}_{i=1}^p, x_i \in  \mathbb{R}^{3} \hspace{0.5em} \forall i \in\{1, \ldots, p\} $ for a part set with $n$ shapes, we form it as a variation matrix
$\bm{X} \in \mathbb{R}^{n \times 3p}$. We compute the mean,$\overline {\mathbf{x}}$, of the variation matrix to normalize it and perform eigen decomposition  $\operatorname{Cov}(\mathbf{X}) = Q \Lambda Q^{\top}$  to 
get the eigenvector matrix $Q$. To keep the latent vector compact but 
 informative, we keep $q$ principle components to create $\hat{Q}$. 
Then we can project each part to the latent space via $\mathbf{Y}= \hat{Q}^{\top} \mathbf{X} \in \mathbb{R}^{n \times q} $. Also, we have 
$E(\mathbf{Y})=\mathbf{0} $ and $\operatorname{Cov}(\mathbf{Y}) =\Lambda $. As a result, we obtain a normalized embedding for each shape part variation, given by \( \mathbf{V} = \frac{\mathbf{Y}}{\sqrt{\Lambda}} \sim \mathcal{N}(\mathbf{0}, \mathbf{I}) \), where \( \mathbf{V} \in \mathbb{R}^q \). Meanwhile, we save the SSM parameters, including point cloud mean $\overline {\mathbf{x}}$, the principle component matrix $\hat{Q}$, and the latent covariance $\Lambda$ of each part set for the shape recovery step.


\noindent \textbf{Part label Semantics}. As the part geometry latent is continuous, we follow~\cite{regol2023diffusing} to use Gaussian Mixture Models to model the semantics of shape part. For each set, we have $m + 1$ models with an additional one for the category of the empty part. Thus, each part has a label Gaussian embedding vector $c \in \mathbb{R}^{m+1}$ to form $ \mathbf{C} \in \mathbb{R}^{n \times (m+1)}$ for the whole part set. If the number of parts for a shape is smaller than $m$, we pad the remaining dimensions with zeros.

\noindent \textbf{Shape Representation}. 
In the computation of part-level SSM, each part is encoded by preserving the most significant principal components of the linear part feature space, resulting in a latent variable for geometry variation, $ z_V \in \mathbb{R}^{q}$. Also, the semantics of the parts are captured in a latent variable, $ z_C \in \mathbb{R}^{m+1}$.
 Thus, the complete embedding of a part is represented as $ z \in \mathbb{R}^{(q+m+1)}$, integrating both geometric and semantic dimensions and the latent for the whole shape is $Z \in \mathbb{R}^{m \times (q+m+1)}$ as shown in Figure.~\ref{fig:overview}.
 In our method, $q$ is 64 and $m$ is 4 for the ShapeNet part set~\cite{yi2016scalable}, which makes this representation more compact than SPAGHETTI~\cite{hertz2022spaghetti}. Meanwhile, the compositional nature and relative poses of parts are also encoded in our shape part latent set implicitly in this way.

 \noindent \textbf{Full Shape Decoding}. For shape decoding from a latent, we reverse the above process. Given a shape latent $Z \in \mathbb{R}^{m \times (q+m+1)}$, consisting of $m$ parts, we first extract $z_V$ and $z_C$ from each part latent. As $z_V$ is a probability vector, the highest probability decides its semantic class. Then we retrieve the saved SSM parameters for this part and decode the part shape as follows:
 \vspace{-1.5mm}
\begin{equation}
x \approx \overline {\mathbf{x}} + \hat{Q} z \sqrt{\Lambda}.
\vspace{-1.5mm}
\end{equation}
Here, the decoder approximates the original shape part \( X \) by linearly adding the latent \( z \) represented shape variations to the mean shape. This method allows us to represent the compositional variability in a compact latent space. After decoding each latent vector from $Z$ and performing a simple concatenation, the whole shape $S$ is recovered.

\subsection{Categorical Part Set Diffusion}
\label{sec:cate_part_set_diffusion}

Building upon the compositional semantic SSM representation, we can encode 3D shape into a set of compact and low-dimensional shape part latents. We propose a categorical diffusion model tailored for learning the distribution of both structure and part-level variations.


 \noindent \textbf{Forward Process.} As demonstrated in Fig.\ref{fig:overview}, we diffuse on the shape part latent set. Following the diffusion process described by~\cite{ho2020denoising}, we systematically add noise to the latent representation ${Z} \in \mathbb{R}^{m \times (q+m+1)}$ in a manner that varies over time to model the progression of the diffusion as equation \ref{eq:forward} shows.
 \begin{equation}
    \label{eq:forward}
     {Z}^{(t)}=\sqrt{\bar{\alpha}^{(T)}} {Z}^{(0)}+\sqrt{1-\bar{\alpha}^{(T)}} \varepsilon
 \end{equation}
 where $ \varepsilon \sim \mathcal{N} (\mathbf{0}, \mathbf{I})$.
 
\noindent \textbf{Latent Denoising Network.} The latent set $Z$ is composed of both geometry and semantics embeddings. To effectively capture the distinct characteristics of each type of embedding, we employ separate encoder and decoders for the geometry and semantics respectively. The two encoders ensure that each aspect of the embedding is processed according to its specific attributes. The encoded outputs are then concatenated to form a unified representation, as illustrated in Fig~\ref{fig:overview}. We employ a series of time-conditioned transformers to serve as the denoiser network for set data.  Within these blocks, attention modules are tasked with learning the intricate structures and relationships among semantic part latents. AdaIN~\cite{perez2018film} layers function as a conditional normalization mechanism that adaptively rescale and shift the normalized activations based on contextual features. 

At inference time, we simultaneously denoise a noise variable sampled from $\mathcal{N}(\mathbf{0}, \mathbf{I})$ to obtain ${Z_V}$ and a padding label to obtain ${Z_C}$. We then transform the denoised ${Z_V}$ into a point cloud using the corresponding SSM derived from the denoised part label ${Z_C}$.

\noindent \textbf{Loss Functions.} We predict both the ground truth data ${Z_0}$ and the semantic labels it contains, the label is also used in identifying the corresponding part's SSM when decoding the predicted shape part latents. In addition, we propose a Kullback–Leibler(KL) divergence~\cite{kullback1951information} regularization loss for the shape part latents, $ {Z_V}$, whose mean and covariance are known in the SSM. We employ this property to effectively improve the performance of the model to learn the structured SSM representations.

Therefore, the total loss $\mathcal{L}$ for the network is a weighted sum of the mean squared error (MSE) loss $\mathcal{L}_{\text{mse}}$, the cross-entropy (CE) loss $\mathcal{L}_{\text{cls}}$, and the KL-divergence loss $\mathcal{L}_{\text{kl}}$. $\lambda \geq 0$ are weights to balance the three loss components:
 \begin{equation}
    \mathcal{L} := \lambda_1 \cdot \mathcal{L}_{\text{mse}} + \lambda_2 \cdot \mathcal{L}_{\text{cls}}+ \lambda_3 \cdot \mathcal{L}_{\text{kl}}
 \end{equation}
\begin{equation}
\label{eq:lmse}
    \mathcal{L}_{\text{mse}} := \mathbb{E}_{t, {Z_V}^{(0)}, \boldsymbol{\epsilon}}\left[\left\|{Z_V}^{(0)} - \hat{{Z_V}}^{(0)}\right\|^2\right]
\end{equation}
\begin{equation}
\label{eq:cls}
    \mathcal{L}_{\text{cls}} := \mathbb{E}_{t, {Z_C}^{(0)}, \boldsymbol{\epsilon}}\left[ - \sum_{c=1}^C {Z_C}^{(0)}(c) \log \hat{{Z_C}}^{(0)}(c)\right]
\end{equation} 
\begin{equation}
\label{eq:kl}
{
    \mathcal{L}_{\text{kl}}({Z_V})=  
    \frac{1}{2} \Bigl(\,\mu_v^2 + \sigma_v^2 \;-\; \ln(\sigma_v^2) \;-\; 1\Bigr)
}
\end{equation}
where $\mu_v$, $\sigma_v^2$ are the mean and variance of ${Z_V}$.


Furthermore, to ensure the network focuses on the relationships of shape part latents, we mask out the padding latent in the $\mathcal{L}_{\text{mse}}$ and $\mathcal{L}_{\text{kl}}$. The $\mathcal{L}_{\text{kl}}$ loss, derived from the inherent properties of our SSM representation, is further evaluated through an ablation study in the unconditional shape generation experiment section.



\subsection{Shape Refinement}
As our shape representation approximated the real data with the principle components for capturing both the overall structure and essential geometry, this inevitably results in generating shapes that lose a certain level of details and the joint areas between parts are a little noisy. We thus integrate a refinement step to enhance the high-frequency details. Therefore, we adapt an additional network to refine the generated point cloud and output the fine-grained shape as a signed distance field (SDF). Finally, we run the Marching Cubes to extract the mesh from it. The refined shapes are presented in Fig.~\ref{fig:unconditional_generation} and ~\ref{fig:single_recons}.

\section{Experiment}
\label{sec:exp}

\begin{table*}[t!]
\centering
\newcolumntype{Y}{>{\centering\arraybackslash}X}
\caption{\textbf{Quantitative comparison of unconditional shape generation.}
MMD-CD scores and MMD-EMD scores are scaled by $10^3$ and $10^2$, respectively. \textbf{The best results are highlighted.} The ablation study results are shown in row 4. All data
normalized individually into $[-1,1]$.
}

\label{tab:quantitative_comparison_of_shape_generation}
\footnotesize
\setlength{\tabcolsep}{0.2em}
\renewcommand{\arraystretch}{1.0}
\definecolor{LightCyan}{rgb}{0.88,1,1}
\definecolor{Gray}{gray}{0.85}
\begin{tabularx}{\linewidth}{>{\centering}m{3.5cm}| Y Y Y Y Y Y | Y Y Y Y Y Y }
  \toprule
  \multirow{3}{*}{Method} & \multicolumn{6}{c|}{Chair} & \multicolumn{6}{c}{Airplane} \\
                & \multicolumn{2}{c}{COV (\%, $\uparrow$)} & \multicolumn{2}{c}{MMD $\downarrow$} & \multicolumn{2}{c|}{1-NNA(\%, $\downarrow$)}  & \multicolumn{2}{c}{COV (\%, $\uparrow$)} & \multicolumn{2}{c}{MMD $\downarrow$} & \multicolumn{2}{c}{1-NNA(\%, $\downarrow$)} \\
                &   CD   &   EMD   &   CD   &   EMD   &   CD   &   EMD   &   CD   &   EMD   &   CD   &   EMD   &   CD   &   EMD \\
  \midrule
  Diffacto~\cite{nakayama2023difffacto}  & 39.97  &	22.86 &	15.79 &	20.93 &	75.95 &	96.78 & 46.97 &	24.10 &	4.69 &	14.15 &	91.42 &	97.72\\
  \midrule
SALAD~\cite{koo2023salad} & 45.13	&45.50	& 13.20	&14.87	&67.40	&\textbf{66.60} & 55.75	& 60.69	& 4.37	& 8.74	& 85.61	& 82.45\\  
  \midrule 

Ours w/o KL  & 48.86	& 47.25 & 13.40  & 16.64	& 73.31 	& 73.32  	 &59.30 & 63.05	& 4.21	& 10.19	& 86.74	& 83.18 \\

\ourMethod{}(Ours) & 49.32	& 48.70 & 13.10  & 15.14	& 70.31 	& 68.75  	 &63.94 & 65.57	& 4.13	& 9.33	& 81.69	& 78.50 \\
 \midrule 
\ourMethod{}(w/ refinement)  & \textbf{52.74}	& \textbf{49.65} & \textbf{11.48}  & \textbf{13.29}	& \textbf{65.31} & 67.53  	 
&\textbf{64.72} & \textbf{66.83}	& \textbf{4.07}	& \textbf{9.11}	& \textbf{79.02}	& \textbf{76.29} \\
 
\bottomrule
\end{tabularx}
\vspace{-1.5mm}
\end{table*}
In this section, we conduct extensive qualitative and
quantitative experiments to demonstrate the efficacy of the proposed \ourMethod{} in learning structure and  variations of 3D shape.
We perform evaluation on the following tasks: unconditional shape generation, single-view 3D reconstruction, shape generation and text-guided generation.

\subsection{Unconditional Shape Generation}

\noindent \textbf{Evaluation Setup.} In this section, we evaluate our \ourMethod{} on the task of unconditional generation. We demonstrate the performance of generating complex structure shape and variations by comparing with SALAD~\cite{koo2023salad} and Diffacto~\cite{nakayama2023difffacto}, which are the state-of-the-art methods on structure-aware shape generation. We follow~\cite{koo2023salad} to use metrics of  minimum matching distance (MMD), coverage (COV), and 1-nearest neighbor accuracy (1-NNA) based on the Chamfer distance for evaluation. We share the same shape part dataset constructed based on ShapeNet~\cite{chang2015shapenet} with~\cite{nakayama2023difffacto} and the part label is from ~\cite{yi2016scalable}. The dataset~\cite{yi2016scalable} provides maximal four part labels for each shape. 
The two standard classes, the chair and airplane, comprise 3,053 and 2,349 training shapes respectively, with corresponding test sets of 704 and 341 shapes respectively. Due to the limited size of the test set, which does not adequately capture the true distribution of shape data, direct evaluation of generated data leads to significantly biased metrics. Following ~\cite{koo2023salad}, we evaluate all retrained models using a test split comprising 1,356 chair shapes and 809 airplane shapes.
For evaluation, we generate 2000 shapes for each class. We save 64 principal components of the shape part SSM in our experiments. Refer to the \textbf{supplementary} for more details of choosing the number of  principal components.

\begin{figure}

\includegraphics[width=0.98\linewidth]{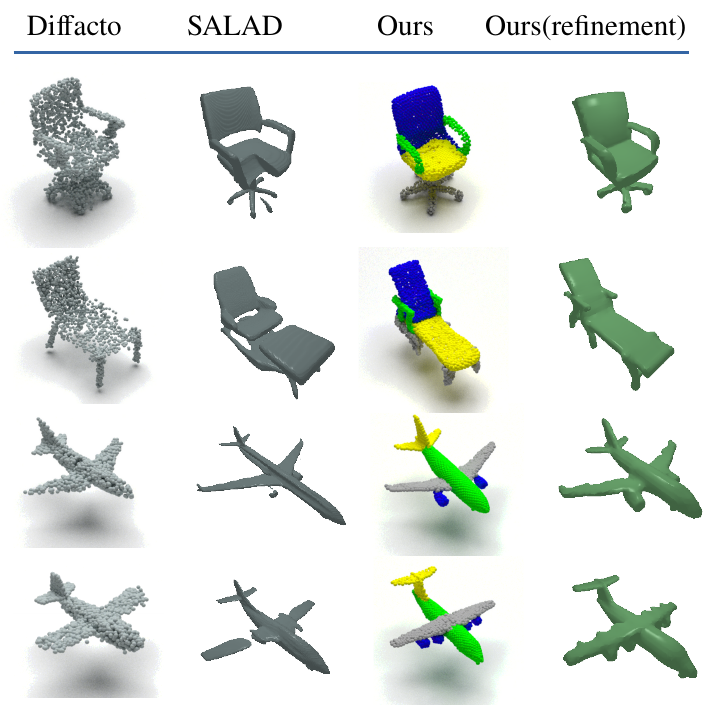}
\caption{\textbf{Qualitative results for unconditional shape generation.} Given a ground truth shape, we retrieve the closest generated shape by evaluating the EMD for each method. Our approach yields superior results in producing complex shape structures and high levels of detail after refinement compared to other methods.}
\label{fig:unconditional_generation}
\vspace{-0.7em}
\end{figure}

\noindent \textbf{Results.} The quantitative and qualitative results, including ablation studies, are summarized in Table~\ref{tab:quantitative_comparison_of_shape_generation} and Figure~\ref{fig:unconditional_generation}. Table~\ref{tab:quantitative_comparison_of_shape_generation} shows our method achieves SotA results or is on par with the baselines. Our method gets significantly better score in the COV metric, indicating our method can generate more diverse and high-quality shapes. In terms of MMD, without refinement, due to the limitation of SSM that only limited principle components are used, our MMD is lower than SALAD~\cite{koo2023salad}, but still outperforms~\cite{nakayama2023difffacto}.
With our high-quality point cloud and the refinement network, our refined mesh quality outperforms other methods.

For qualitative comparisons in Figure~\ref{fig:unconditional_generation}, we retrieve generated shapes using the same query as the ground truth shape and compare them.

\subsection{Single-view Reconstruction}
\label{sec:single_view_recons}

\begin{figure}
\centering
\includegraphics[width=0.99\linewidth]{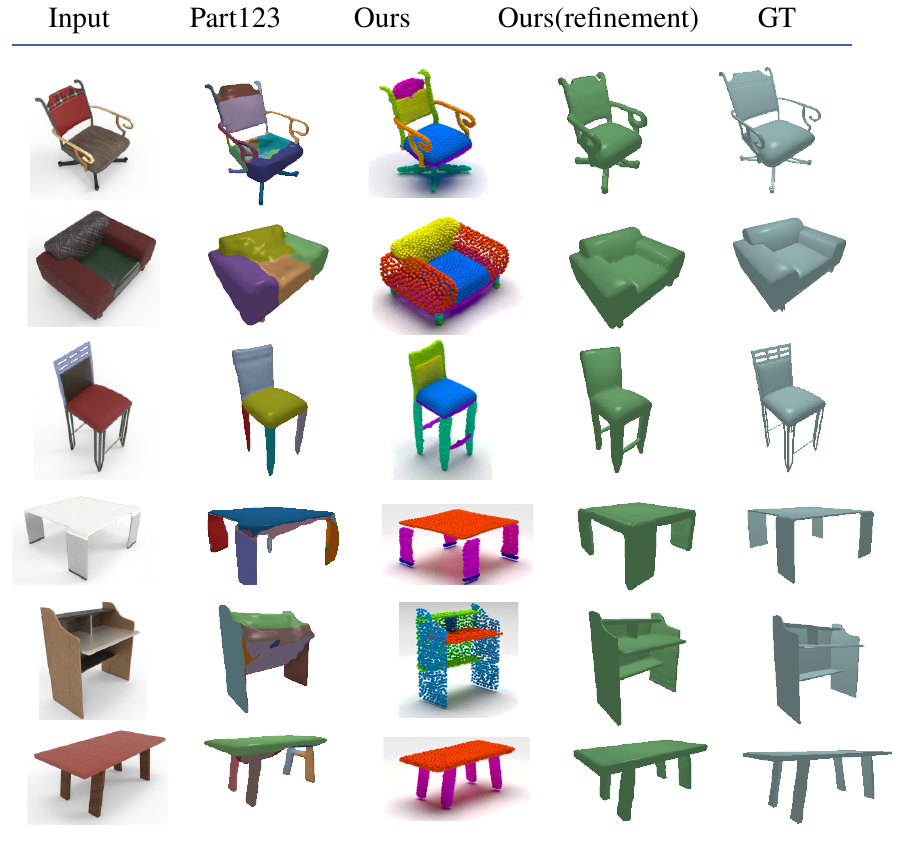}
\caption{\textbf{Part-level single-view reconstruction.} We qualitatively compare with Part123~\cite{liu2024Part123} on the 3DCoMPaT++ dataset~\cite{slim_3dcompatplus_2023}. Given a single view image, our method produces 3D shape that is more consistent with the image even for the complex structures.
}
\label{fig:single_recons}
\vspace{-0.7em}
\end{figure}

We evaluate the single-view reconstruction task on a fine-grained part-level dataset 3DCoMPaT++ ~\cite{slim_3dcompatplus_2023} to demonstrate our method in learning the compositional nature of complex shape structure. The selected chair and table sets have 23 and 22 labels respectively, and each shape may have a maximum of 11 different parts, making it more challenging than ShapeNet part dataset~\cite{yi2016scalable}. We collect the part-level 3D shapes from the dataset and split 80\% of the data for training and 20\% for testing. We rendered 8 fixed-view images with a resolution of 256×256 for each shape.

To condition the model on an image, we employ cross-attention to incorporate the image features into the shape latent representation. We compare our method with the SotA part-aware work Part123~\cite{liu2024Part123}, which leverages an image diffusion model and SAM~\cite{kirillov2023segment} to generate multiview segmentations of an object from a single-view image and the part-aware reconstruction algorithm to reconstruct the 3D shape. As there is no evaluation result on 3DCoMPaT++~\cite{slim_3dcompatplus_2023} of Part123~\cite{liu2024Part123}, we fine-tune their model on 3DCoMPaT++~\cite{slim_3dcompatplus_2023}. We adopt two commonly used metrics, Chamfer Distances and Volume IoU, to measure the difference between the reconstructed and ground truth shapes.

\noindent \textbf{Results.} 
The quantitative results in Table.~\ref{tab:recon} and qualitative results in Figure~\ref{fig:single_recons} show that our method outperforms Part123~\cite{liu2024Part123}  and that our method can generate point cloud and refined mesh of high quality. As shown in Figure~\ref{fig:single_recons}, Part123 tends to produce over-smoothed shapes and struggles with complex structure, whereas our method consistently delivers high-quality, fine-grained part-level 3D reconstructions, even for images with intricate typologies.




\begin{table}
    \centering

     \caption{\textbf{Quantitative comparison of single-view reconstruction.}
     We compare with part123~\cite{liu2024Part123}. We report Chamfer Distance and Volume IoU on the 3DCoMPaT++ dataset~\cite{slim_3dcompatplus_2023}.}
     
    \begin{tabular}{ccc}
       \toprule
       Method  & Chamfer Dist.$\downarrow$ & Volume IoU$\uparrow$ \\
       \midrule
       Part123
       & 0.0416 &  0.308 \\
       \ourMethod (Ours)    
       &  \textbf{0.0221}  &  \textbf{0.412}   \\
       \bottomrule
    \end{tabular}
   
    \label{tab:recon}
    \vspace{-10pt}
\end{table}

\subsection{Text-conditioned Shape Generation}
\begin{table}[t!]
\centering
\newcolumntype{Y}{>{\centering\arraybackslash}X}
\caption{\textbf{Quantitative comparison of text-guided generation.} Overall, our method achieves better performance than SALAD. Specifically, it improves FID by a large margin.}
\scriptsize
\setlength{\tabcolsep}{0.2em}
\renewcommand{\arraystretch}{1.0}
\begin{tabularx}{\linewidth}{>{\centering}m{2.1cm}| Y | Y | Y | Y}
\toprule
Methods & IoU$\uparrow$  & CD$\downarrow$ & F-score$\uparrow$ &FID$\downarrow$ \\
\midrule
SALAD & 14.02 & 1.33 & \textbf{13.58} & 1.25 \\ 
\ourMethod (Ours)  & \textbf{14.56} & \textbf{1.27} & 12.7 & \textbf{ 0.83} \\ 
\bottomrule
\end{tabularx}
\label{tbl:lang_quantitative_result}
\vspace{-1.5mm}
\end{table}

To evaluate the performance of text-guided shape generation, we conduct experiments on the ShapeNet dataset with textural description from  Text2Shape~\cite{chen2019text2shape}. We compare our method with the SotA part-level method SALAD~\cite{koo2023salad}. To assess the generation quality, we adopt Intersection over Union (IoU), Chamfer Distance (CD), F-score, and Fréchet Inception Distance (FID) as evaluation metrics. All the metrics are computed between the generated and GT shapes.

To integrate textual information, we extract language-derived features using BERT~\cite{devlin2019bert} and feed them into an AdaIN~\cite{perez2018film} adapter which introduces a gating mechanism. This mechanism dynamically modulates each element in the shape latent set, adapting the shape features based on the contextual cues provided by the text.
During the inference, we follow the classifier-free guidance ~\cite{ho2022classifier} scheme.

\begin{figure*}[ht]
\includegraphics[width=0.92\linewidth]{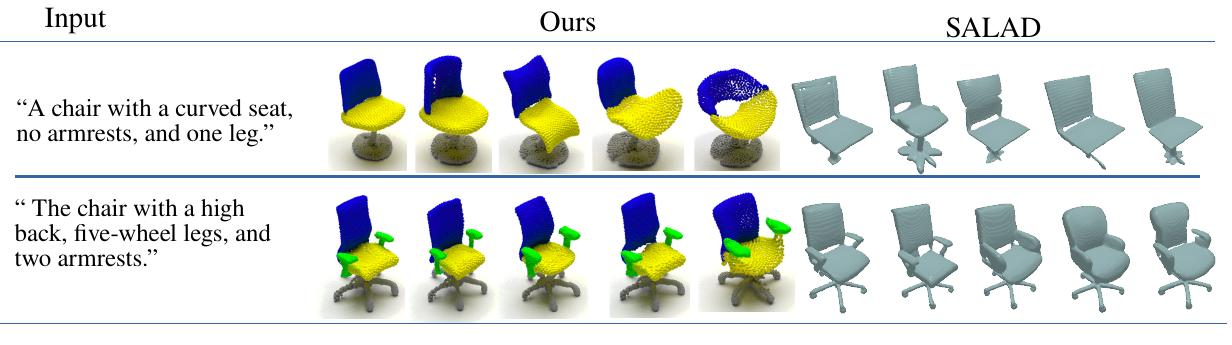}
\caption{\textbf{Qualitative comparison of text-to-shape.} \ourMethod{} generates more diverse, high-quality results that align with given texts. }
\label{fig:textCond_quali}
\vspace{-1.5em}
\end{figure*}

\noindent \textbf{Results.}
As demonstrated in Table~\ref{tbl:lang_quantitative_result}, our method achieves superior results, particularly in FID. This improvement suggests that our SSM latent set representation more effectively aligns the generated 3D shapes with corresponding text features.
As shown in Fig.~\ref{fig:textCond_quali}, our approach consistently outperforms SALAD~\cite{koo2023salad} by
generating high-quality shapes that closely align with textual descriptions, demonstrating its capability to capture various structures and intricate geometry.

\section{Application}
\label{sec:exp}

\subsection{Cascaded Part Completion}

\begin{figure}
\includegraphics[width=0.96\linewidth]{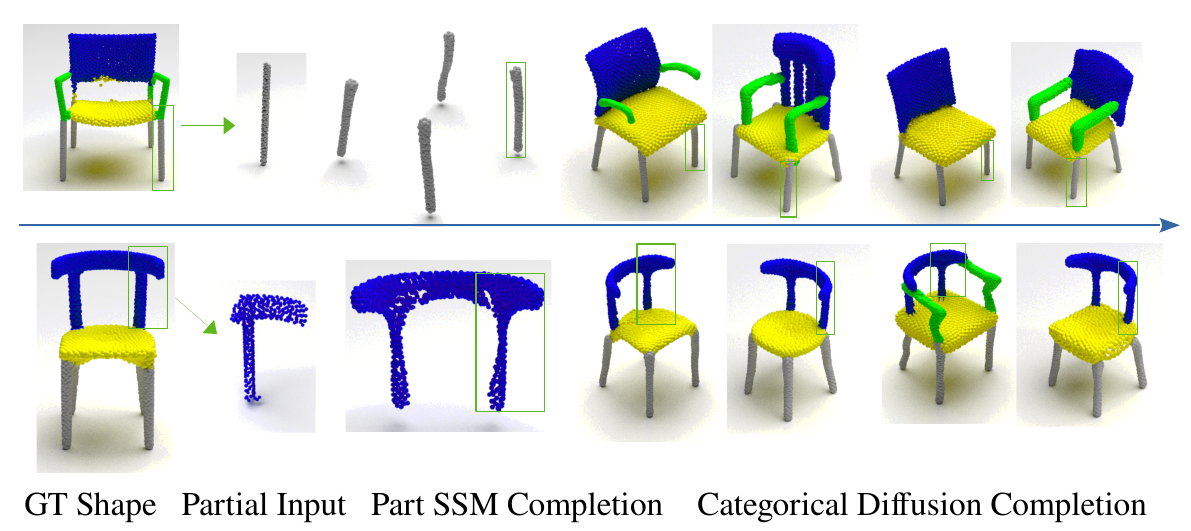}
\caption{\textbf{Cascaded part completion.} Our method generates complete shapes from a sub-part input. It derives the complete part from a sub-part using corresponding SSM and applying diffusion denoising to produce a variety of plausible whole shapes. }
\label{fig:part_completion}
\vspace{-0.7em}
\end{figure}

In this section, we introduce a novel part completion using our part-level SSMs and trained unconditional categorical diffusion model. For completion using diffusion models, the guided reverse process  \cite{meng2021sdedit} is commonly employed. With this approach, methods like SALAD~\cite{koo2023salad} and Neural Wavelet ~\cite{hui2022neural} mask out the part inputs in the representation and then regenerate the full shape. However, due to the implicit nature of their representations, these methods cannot explicitly mask the completed parts.
Consequently, they are unable to achieve fine-grained part-level shape completion. Like previous methods such as ShapeFormer~\cite{yan2022shapeformer}, our approach preserves the given part.

Our method can generate complete and diverse shapes only from a sub-part input, which is realised by the part SMM completion to recover the full part and the diffusion completion to complete the whole shape. As shown in Figure~\ref{fig:part_completion}, our cascaded part completion method can first derive the complete part from a sub-part and then generate the whole shape from the part, i.e. from one leg to all chair legs and to fully complete and diverse shapes. 

Given a partial point cloud of a complete part along with its part label, we first employ a KNN search to match it against the mean shape of the SSMs to identify the corresponding SSM of the observed point cloud.
Then we optimize a part linear SSM parameters $\mathbf{z_v}$ to fit the observed points in a least-squares way to output the results of SSM parameters and SSM completed part shape. 
After optimizing $\mathbf{z_V}$, we pad it and its label into a shape-level latent set by incorporating noise drawn from $\mathcal{N}(\mathbf{0}, \mathbf{I})$ along with a predefined padding label. During the denoising phase of our categorical diffusion model, the observed latent and its label remain fixed while the remaining components of the shape latent set are denoised.

\subsection{Part Mixing and Refinement}
In addition, our method supports refining the mixed shape part using learned structure and geometry priors. As shown in Figure~\ref{fig:teaser} (b), the newly added armrest does not align with the topology of the original shape. To refine the incoherent part with our diffusion prior, we denoise the first half of its SSM latent vector dimensions, while the remaining dimensions of the latent vector and the other parts of the object are fixed during the denoising process. 
The results indicate that targeted denoising on the principal dimensions of our latent vector, combined with an effective global geometry prior, significantly refines the mixed components. Refer to the \textbf{supplementary material} for more details.

\subsection{Texture Mapping}

Since we model shapes with 3D correspondences in SSM, our generated part-level point cloud enables semantically consistent texture mapping across the refined meshes, as demonstrated in Figure~\ref{fig:teaser} (a).

\subsection{Shape Editing}

Our approach to controllable generation supports a broad array of shape-editing applications. By representing shapes with part-level SSM latents, each part is modeled explicitly, enabling our method to perform simple but effective shape manipulation. Adding a new part to an existing shape can be achieved by adding a new valid part latent vector to the shape latent and part replacement can be done by the latent interpolation as shown in Figure~\ref{fig:teaser} (b).
Figure~\ref{fig:teaser} (c) shows that our method also enables interpolation between the targeted part from two generated shapes while keeping all other parts fixed. These shape editing abilities further extend the capability of our method to generate controllable and diverse shapes.



\section{Conclusion}
\label{sec:conclusion}
This paper introduces PRISM, a novel probabilistic framework that integrates Statistical Shape Models with categorical diffusion to address the fundamental challenge of compositional 3D shape modeling. By capturing both part-level geometric variations and valid compositional patterns, our approach overcomes key limitations of existing methods in representing complex real-world objects with varying topological structures. Our comprehensive experimental evaluation demonstrates PRISM's superior performance in unconditional generation, single-view reconstruction, and text-guided generation tasks, particularly for objects with varying part configurations. The integration of part-wise SSMs with structured categorical diffusion enables fine-grained control over shape manipulation while ensuring statistical validity, advancing the state-of-the-art in topologically-aware 3D shape generation and manipulation.

{
    \small
    \bibliographystyle{ieeenat_fullname}
    \bibliography{main}
}



\end{document}